\documentclass{article}
\usepackage[a4paper, margin=3cm]{geometry}
\usepackage[utf8]{inputenc}
\usepackage[colorlinks,allcolors=blue]{hyperref}
\usepackage[noblocks]{authblk}
\usepackage{graphicx}
\usepackage{booktabs}
\usepackage{amssymb,amsmath,amsfonts}
\usepackage{dsfont}
\usepackage[figuresleft]{rotating}
\usepackage{listings}
\newcommand{\code}[1]{\lstinline[basicstyle=\normalsize\ttfamily]{#1}}

\usepackage{xcolor}

\definecolor{codeback}{rgb}{0.99,0.99,0.99}
\definecolor{codegreen}{rgb}{0.1,0.5,0}
\definecolor{codegray}{rgb}{0.5,0.5,0.5}
\definecolor{codepurple}{rgb}{0.3,0,0.6}
\definecolor{codeblue}{rgb}{0.0,0,0.7}
\definecolor{codered}{rgb}{0.4,0.0,0.0}

\lstdefinestyle{paradiseo}{
    xleftmargin=10pt,
    backgroundcolor=\color{codeback},   
    commentstyle=\rmfamily\color{codegreen},
    keywordstyle=\color{codered},
    morekeywords={size_t},
    numberstyle=\tiny\color{codegray},
    stringstyle=\color{codepurple},
    basicstyle=\footnotesize\ttfamily,
    breakatwhitespace=false,         
    breaklines=true,                 
    captionpos=b,                    
    keepspaces=true,                 
    numbers=left,                    
    numbersep=5pt,                  
    showspaces=false,                
    showstringspaces=false,
    showtabs=false,                  
    tabsize=2
}

\usepackage{enumitem}
\setlist[description]{leftmargin=\parindent,labelindent=\parindent} % indent description lists

\newcommand{\Ind}[1]{\ensuremath{\mathds{1}(#1)}}

\renewcommand{\vec}[1]{\mathbf{#1}}
\newcommand{\reals}[1]{\mathbb{R}}
\newcommand{\ints}[1]{\mathbb{N}}
\newcommand{\weakdom}{\preceq} % Carola prefers \leq instead of \preceq
\newcommand{\ECDF}{\ensuremath{\widehat{F}}}
\newcommand{\EAF}{\ensuremath{\widehat{G}}}

\newcommand{\IOH}[1]   {IOH{\lower 0ex \hbox{\smaller{\textmd{#1}}}}}
\newcommand{\name}[1]   {{\fontfamily{lmss}\selectfont {#1}}}

\usepackage{relsize} % \larger and \smaller
\def\Cpp{%
    C\kern-.1em\raise.30ex\hbox{\smaller{++}}%
\spacefactor1000 }

\title{Extensible Logging and\\
Empirical Attainment Function\\
for \IOH{experimenter}}
\author{Johann Dreo}
\affil{Systems Biology Group, Computational Biology Department, Institut Pasteur, Université de Paris, France}
\author{Manuel L\'opez-Ib\'a\~nez}
\affil{School of Computer Science and Engineering, University of M{\'a}laga, M{\'a}laga, Spain}
\date{30 September 2021}

\begin{document}

\maketitle

\begin{abstract}
In order to allow for large-scale, landscape-aware, per-instance algorithm selection,
a benchmarking platform software is key.
\IOH{experimenter} provides a large set of synthetic problems, a logging system and a fast implementation.

In this work, we refactor \IOH{experimenter}'s logging system, in order to make it more extensible and modular.
Using this new system, we implement a new logger, which aim at computing performance metrics of an algorithm across a benchmark.
The logger computes the most generic view on an anytime stochastic heuristic performances,
in the form of the Empirical Attainment Function (EAF).
We also provide some common statistics on the EAF and its discrete counterpart, the Empirical Attainment Histogram.

Our work has eventually been merged in the \IOH{experimenter} codebase.
\end{abstract}

\section{Introduction}

\IOH{experimenter}~\cite{IOHprofiler} is a framework for the benchmarking of anytime stochastic heuristics for optimization problems
that holds together a set of benchmarks and an advanced logging system.
It provides Pseudo-Boolean Optimization problem generators (PBO) and the Black-Box Optimization Benchmark (BBOB).
Its core is implemented in \Cpp, which allow for fast computations, a crucial feature for benchmarking on synthetic problems.

One of the most notable feature of \IOH{experimenter} is its logging system.
A so-called {\em Logger} target an automated export of the algorithms' behaviour in a standardized format,
which can be seamlessly imported in the \IOH{analyzer} for further study.
It most notably allows for tracking the states of the algorithms' parameters, a unique feature on the market of benchmarking platforms.

%\section{Legacy \IOH{experimenter} Logging System}

The legacy logging system was implemented as a single class, exposing a set of hard-coded parameters
which controlled when logging event occurred,
what parameters were to be watched
and how they would be stored in a set of CSV files.

The limitations of the legacy system were:
\begin{itemize}
    \setlength{\itemsep}{1pt}
    \setlength{\parskip}{0pt}
    \setlength{\parsep}{0pt}
    \item The logging event list was not extensible.
    \item Only plain floating-point parameters could be logged.
    \item Parameters had to be explicitly sent by the calling solver and should always exists.
    \item It was only possible to log into the \IOH{analyzer} format, in a set of CSV text files.
\end{itemize}

We show in section~\ref{sec:archi} how we remove those limitations.

The new logging system is also more extensible and allows for easiest implementation of new loggers.
This feature allowed us to add a new logger which targets immediate performance estimation,
so as to help for automated algorithm selection.
By binding the optimized problems and the performance estimation of a solver in a single binary,
we allow for very fast benchmarking runs.

Having fast benchmarking is a crucial feature for allowing per-instance algorithm configuration.
As shown in \cite{a_AzizAlaoui2021}, having the performance estimation directly bound to the solver's binary
allows for configuration budget that are an order of magnitude larger than other approaches.

In a previous work, we implemented an \IOH{experimenter} logger which efficiently computes the discrete version of the empirical attainment function on a linear scale.
In this work, we refactor this logger, adding log scales, and add a logger computing the empirical attainment function itself, as explained in section~\ref{sec:EAF}.
We also refactored the debugging tools within \IOH{experimenter}, by using the \name{Clutchlog}\footnote{\url{https://nojhan.github.io/clutchlog/}} project, which allow for fine-grained management of debug messages.

\section{Logging System Architecture}\label{sec:archi}

\begin{sidewaysfigure}
    \centering
    \includegraphics[width=1.0\textwidth]{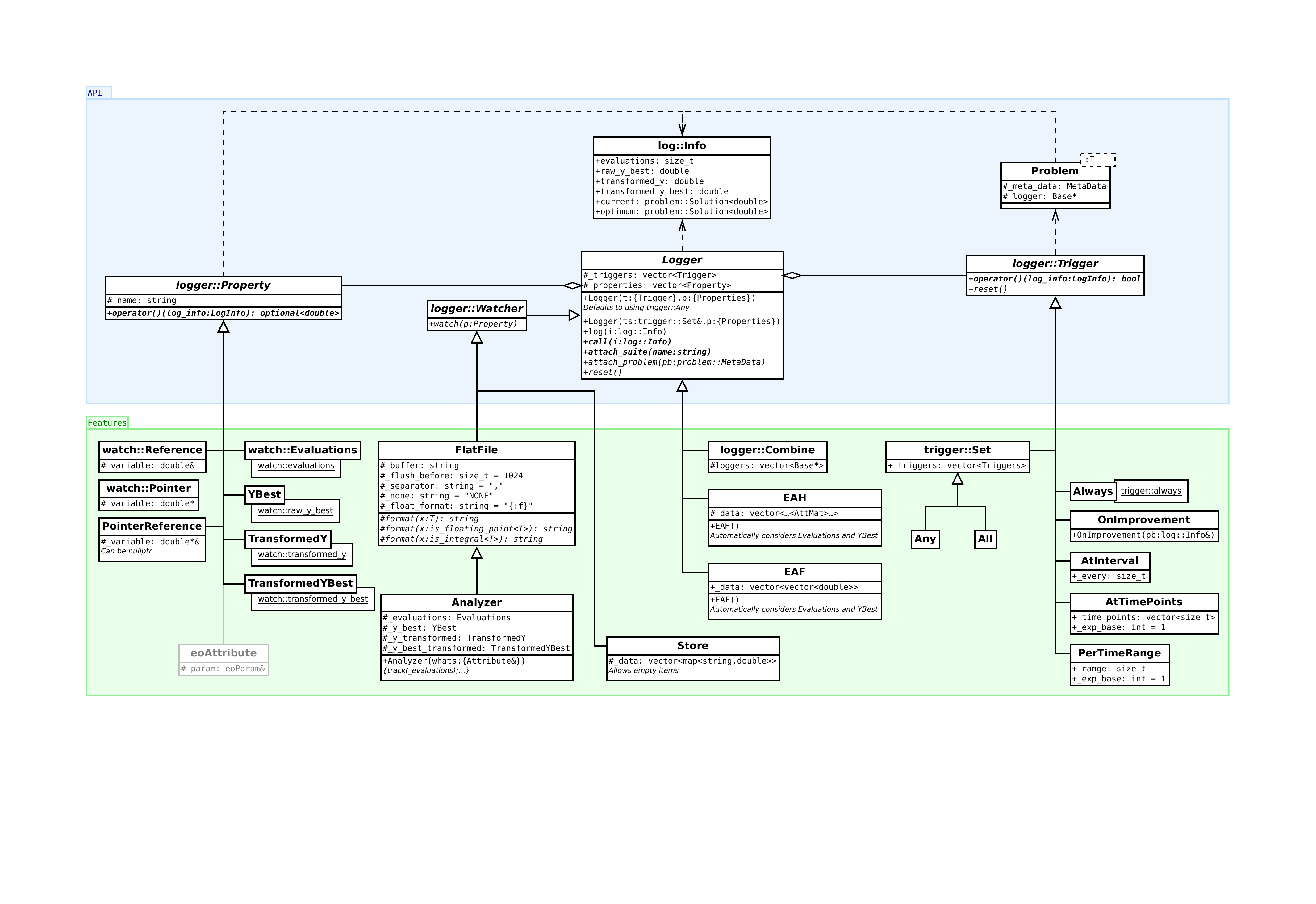}
    \caption{UML class diagram of the architecture that we implemented for the logging system.}
    \label{fig:archi} 
\end{sidewaysfigure}

Instead of a monolithic Logger, we design a set of modular classes, as shown on Figure~\ref{fig:archi}.

A \name{Logger} is thus most notably composed as two objects:
\begin{description}
    \item[\name{Trigger}:] given the context at call, returns true if the call event is to be logged (cf. Section~\ref{sec:triggers}),
    \item[\name{Property}:] gives access to what value should be logged (cf. Section~\ref{sec:properties}),
\end{description}
    
Subclassing from the \code{Logger} interface allow to design a fully configurable object, with user-defined Triggers and Properties.
For instance, the \code{Watcher} class allows the user to specify which Properties to watch.
The standardized logging system for \IOH{analyzer} is itself defined as a sub-class of Watcher.

We add a Logger object that allow for combining several loggers into a single one: \code{Combine}.
For instance, this wrapper is useful if one want to computes a global performance and save the whole run.
This system allow to assemble Loggers at run time.

\subsection{Triggers}\label{sec:triggers}

The Triggers are functors\footnote{Objects that are callable as functions.} which can be called on the \code{LogInfo} context and returns a boolean, stating if a log event should be triggered.

The sets of available Triggers is straightforward and matches the previous features, they allow to log an event:
\begin{description}
    \item[\name{Always}:] triggers an event at every objective function call,
    \item[\name{OnImprovement}:] if the given solution has a better objective function value than the previous best one,
    \item[\name{At}:] at the given set of objective function call indices,
    \item[\name{Each}:] at regular intervals,
    \item[\name{During}:] during the given sets of objective function call intervals,
\end{description}

Instead of managing those Triggers all at once, we allow to aggregate them with triggers that behaves like logical operators:
\begin{description}
    \item[\name{Any}:] trigger an event if {\em any} of the managed Trigger instances is triggered,
    \item[\name{All}:] trigger an event if {\em all} of the managed Trigger instances is triggered.
\end{description}

\newpage 

\begin{lstlisting}[language=C++,label=lst:triggers,
caption={Excerpt of the code used to test the behaviour of the \code{OnImprovement} Trigger.}
]
problem::MetaData pb(0,0,"fake",2);
logger::Info i;
i.transformed_y = 9999;

trigger::OnImprovement t;

// First call.
EXPECT_TRUE(t(i,pb)); // Better than infinity.

i.transformed_y = 100;
EXPECT_TRUE(t(i,pb)); // Improvement.
EXPECT_FALSE(t(i,pb)); // Strict inequality.

// Improvement.
i.transformed_y = 10;
EXPECT_TRUE(t(i,pb)); // Improvement.
EXPECT_FALSE(t(i,pb)); // Strict inequality.

// Unsuspected increasing.
i.transformed_y = 99;
EXPECT_FALSE(t(i,pb)); // Not improving.

// Internal state stability.
i.transformed_y = 11;
EXPECT_FALSE(t(i,pb)); // Still worst than internal test.

i.transformed_y = 9;
EXPECT_TRUE(t(i,pb)); // Better than internal state.

t.reset();
i.transformed_y = 99;
EXPECT_TRUE(t(i,pb)); // Better than infinity.
\end{lstlisting}

\subsection{Properties}\label{sec:properties}

A Property is a way to access a value of interest, which can be either a state of the \IOH{experimenter} context or an external variable of interest.

The set of available Properties accessing the context matches the previous features:
\begin{description}
    \item[\name{Evaluations}:] the number of calls to the objective function so far,
    \item[\name{YBest}:] the best objective function value found so far,
    \item[\name{TransformedY}:] the current objective function value, with invariant-testing transformation(s) applied,
    \item[\name{TransformedYBest}:] the best transformed value found so far.
\end{description}

We add a set of classes to help accessing several kind of external variables, which targets any algorithm parameter:
\begin{description}
    \item[\name{Reference}:] capture a reference to a variable,
    \item[\name{Pointer}:] capture a pointer to a variable,
    \item[\name{PointerReference}:] capture a pointer to a reference to a variable.
\end{description}

The two last classes allow for indicating if the Property is existing (or not) in the current context.
This is useful for dynamic algorithm configuration, for which a given parameter may exists only for some variant.

For instance, if one is logging the state of an algorithm that switch its mutation operator, some of those operators may have various parameters.
Those parameters would be available only if this specific operator is actually instantiated.

The low-level interface uses \Cpp's standard library's \code{std::optional} feature to indicate if this Property is available in this context.
Pointer-based capturing classes are provided as a generic way to interact with any code, as it is sufficient to set the pointer to \code{nullptr}
to indicate that a Property is disabled in the current context.

\begin{lstlisting}[language=C++,label=lst:properties,
caption={Excerpt of code used to test the behaviour of some context Properties.}
]
using namespace ioh;

suite::BBOB suite({1, 2}, {1, 2}, {3, 10}); // problems, instances, dimensions

// Properties that watch my_attribute:
double my_attribute = 0;
watch::Reference attr("Att_reference", my_attribute);
watch::Pointer   attp("Att_pointer"  ,&my_attribute);
double* p_transient_att = nullptr;
watch::PointerReference attpr("Att_PtrRef", p_transient_att);

// Instantiate some properties:
trigger::Always always;
watch::Evaluations evaluations;
watch::RawYBest raw_y_best;
watch::TransformedY transformed_y;
watch::TransformedYBest transformed_y_best;

logger::Store logger({always},
    // Attach those properties to the logger:
    {evaluations, raw_y_best, transformed_y, transformed_y_best, attr, attp, attpr});
suite.attach_logger(logger);

/* [Call the problem function...] */

// One can select which metadata layer to consider:
logger::Store::Cursor first_eval(suite.name(), /*pb*/1, /*dim*/3, /*ins*/1, /*run*/0, /*eval*/0);

// And recover a property value at this event:
auto evals = logger.at(first_eval, evaluations);
\end{lstlisting}

\subsection{Debugging Messages}

We refactor the legacy debug messaging system, replacing it with a set of macros using the \name{Clutchlog} library.
This allows to (de)clutch messages for a given: log level, source code location or call stack depth.
Additionally, the unit tests interface use this features, which can be used from the command line interface.

\begin{lstlisting}[language=Bash,label=lst:clutchlog,
caption={Examples of debugging messages tuning, upper part shows messages up to the \code{Note} level for \code{eaf.hpp} file,
lower part shows messages up to the \code{Debug} level for any file containing the \code{eaf} word.}
]
$ ./test_eaf --gtest_filter=*eaf_logger Note ".*eaf\.hpp"
[...]
[test_eaf] N:>>>>>>>>>>>>>>>> Attach to: pb=1, dim=10, ins=1, run=0 attach_problem @ eaf.hpp:245
[test_eaf] N:>>>>>>>>>>>>> reset                                             reset @ eaf.hpp:289
[test_eaf] N:>>>>>>>>>>>>> Attach to: pb=1, dim=10, ins=1, run=1    attach_problem @ eaf.hpp:245
[test_eaf] N:>>>>>>>>>>>>> reset                                             reset @ eaf.hpp:289
[...]
[test_eaf] N:>>>>>>>>>>>>> Attach to: pb=2, dim=30, ins=2, run=1    attach_problem @ eaf.hpp:245
[test_eaf] N:>>>>>>>>>>>>> reset                                             reset @ eaf.hpp:289
[test_eaf] N:>>>>>>>>>>>>> Attach to: pb=2, dim=30, ins=2, run=2    attach_problem @ eaf.hpp:245

$ ./test_eaf --gtest_filter=*eaf_logger Debug ".*eaf.*"
[...]
[test_eaf] D:>>>>>>>>>>> Attach suite BBOB to logger                   TestBody @ test_eaf.cpp:16
[test_eaf] N:>>>>>>>>>>>>>>>> Attach to: pb=1, dim=10, ins=1, run=0 attach_problem @ eaf.hpp:245
[test_eaf] I:>>>>>>>>>>> pb:1, dim:10, ins:1                           TestBody @ test_eaf.cpp:20
[test_eaf] I:>>>>>>>>>>> > run:0                                       TestBody @ test_eaf.cpp:22
[test_eaf] D:>>>>>>>>>>>>>> EAF called after improvement                   call @ eaf.hpp:265
[test_eaf] N:>>>>>>>>>>>>> reset                                          reset @ eaf.hpp:289
[test_eaf] N:>>>>>>>>>>>>> Attach to: pb=1, dim=10, ins=1, run=1 attach_problem @ eaf.hpp:245
[...]
test_eaf] I:>>>>>>>>>>> Result fronts:                     TestBody @ test_eaf.cpp:34
[test_eaf] I:>>>>>>>>>>> > Front size=1                    TestBody @ test_eaf.cpp:45
[test_eaf] I:>>>>>>>>>>> > Front size=1                    TestBody @ test_eaf.cpp:45
[test_eaf] I:>>>>>>>>>>> > Front size=2                    TestBody @ test_eaf.cpp:45
[...]
[test_eaf] I:>>>>>>>>>>> > Front size=4                    TestBody @ test_eaf.cpp:45
\end{lstlisting}

\section{Empirical Attainment Loggers}\label{sec:EAF}

\begin{figure}[htbp]
    \centering
    \includegraphics[width=1.0\textwidth]{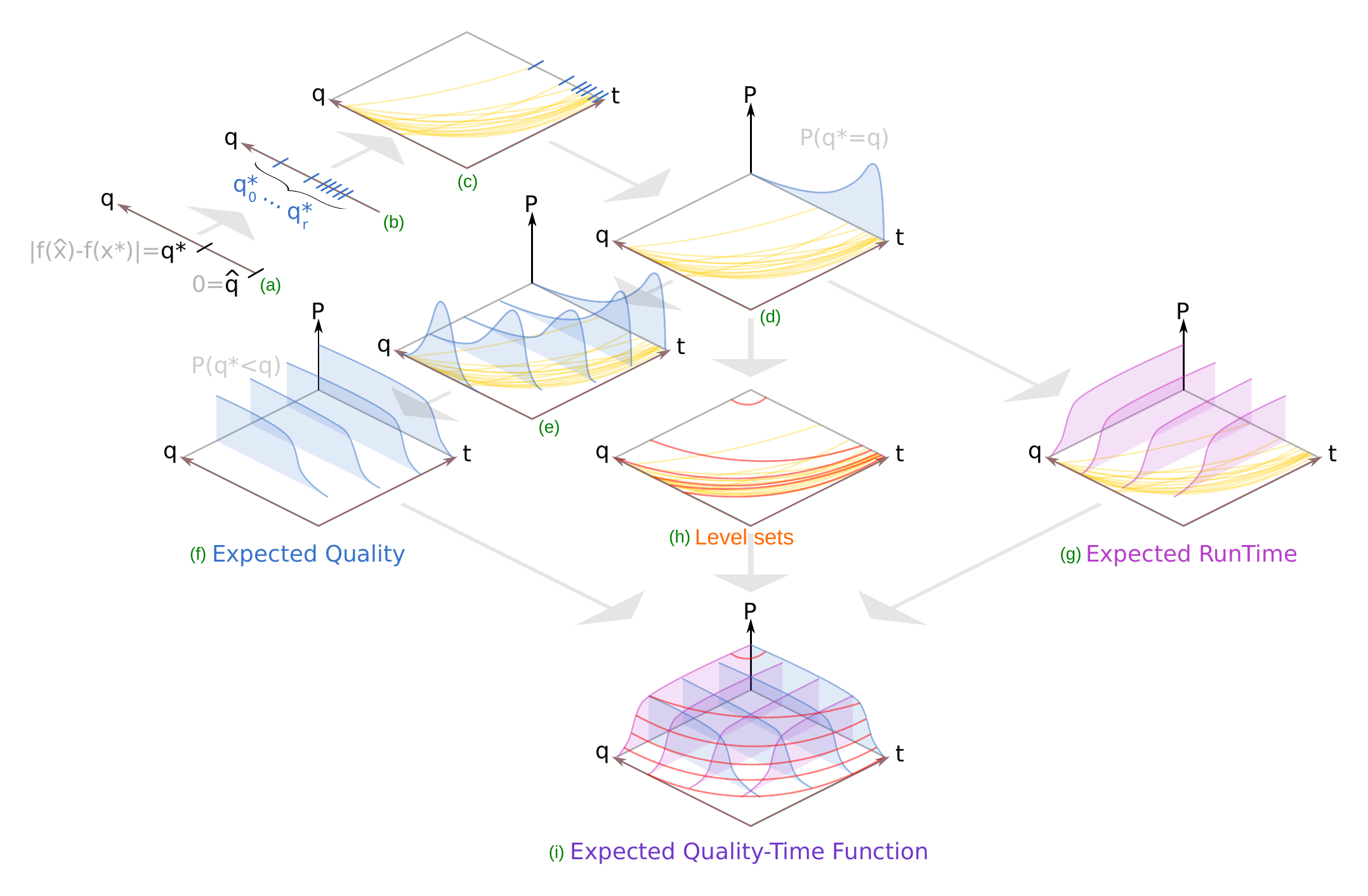}
    \caption{
        A classical way of looking at the performance is to look at the objective function value as the error (``quality'' $q^*$) to a known bound $\hat{q}$, after a given ``time'' budget (a).
        For randomized algorithm, a statistic over the qualities output by several runs should be considered (b).
        But stochastic searches are anytime, hence showing monotonic ``attainment trajectories'' of best-so-far values along the time axis (c).
        A classical setting is to compute a performance statistic over the end distribution (d).
        From those two quality and time axis, one can consider several distribution along the time one (e), most generally in the form of cumulative distribution (f) showing the expected quality for given budgets.
        Conversely, one can consider the expected run time for given qualities (g), which is one of the most popular setting.
        Additionally, one can consider 2D first-order moments of the 2D quality/time distribution of trajectories (h).
        This forms a 2D distribution, called the Empirical Attainment Function (i).
    }
    \label{fig:QTEAF}
\end{figure}

In order to allow for large-scale algorithm configuration, we aim at having a performance evaluation within the solver executable, so as to ease its interfacing with automated configurators.
\IOH{experimenter}, providing both the benchmark problems and a logging system, is a good candidate to host a performance logger.
However, there is many ways of looking at the performance of stochastic heuristics.
In this work, we wanted to implement the most generic one, so as to allow the user to use various performance metrics.
Figure~\ref{fig:QTEAF} shows why we think that the Empirical Attainment Function (EAF) is the most generic choice in the case of synthetic benchmarking.

Section~\ref{sec:ECDF} introduces how the EAF can be viewed as the Empirical Cumulative Distribution Function (ECDF) of 2D objective function values' {\em attainment trajectories},
and how it can be discretized into an Empirical Attainment {\em Histogram} (EAH).

\subsection{Empirical Cumulative Distribution Functions}\label{sec:ECDF}

This section introduces the ECDF of points and then generalize this notion to attainment trajectories, up to the EAF and EAH.
Table~\ref{tab:notation} introduces the notation used in this section.

\begin{table}[htbp]
\caption{Notation.}
\label{tab:notation}
\centering
\begin{tabular}{rl}
  \toprule
  $i, n \in \mathbb{N}$ & integers \\
  $x, y \in \mathbb{R}$ & scalar real values\\
  $\vec{x}, \vec{z} \in \mathbb{R}^2$ & points (vectors) in 2D space\\
  $\mathcal{X}, \mathcal{Y} $ & scalar real random variables\\
  $Z, L \subset \mathbb{R}^2$ & sets of points\\
  $A = \{Z_1,\dotsc,Z_n\}$ & collection, i.e., a multiset of sets\\
  $\Ind{\cdot}$ & Indicator function that returns $\{0,1\}$\\
  $\vec{z}_1 \weakdom \vec{z}_2$ & point $\vec{z}_1$ weakly dominates point $\vec{z}_2$\\
  $Z \weakdom \vec{z}$ & at least one point $\vec{z} \in Z$ weakly dominates point $\vec{z}$\\
  
  \bottomrule
  \end{tabular}
\end{table}

\subsubsection{Of points ---~ECDF}

Let $(\mathcal{X}_1, \dotsc, \mathcal{X}_n)$ be independent, identically distributed real random variables with the common cumulative distribution function $F(x)$. Then, the Empirical Cumulative Distribution Function (ECDF) is classicaly~\cite[p. 265]{van2000asymptotic}\footnote{\url{https://archive.org/details/asymptoticstatis00vaar_017/page/n281/mode/2up}} defined as:
\begin{equation}\label{eq:ECDF_uni}
  \ECDF^n(x) = \frac{1}{n} \sum_{i=1}^n \Ind{x_i \leq x}
\end{equation}
where $\Ind{\cdot}$ is the indicator function that returns 0 or 1 and $x_i \sim \mathcal{X}_i$.

For a pair of random variables $\mathcal{Q}$, $\mathcal{T}$, the joint ECDF is given by:
\begin{equation}\label{eq:ECDF_bi}
  \ECDF^n(q,t) = \frac{1}{n} \sum_{i=1}^n \Ind{q_i \leq q \land  t_i \leq t}
\end{equation}
where $q_i$ and $t_i$ are samples from random variables $\mathcal{Q}$ and $\mathcal{T}$, respectively.

Let $\vec{a} \preceq \vec{b} \iff \vec{a}^d \leq \vec{b}^d \ \forall d \in \ints{}^+$ denote the weak dominance of a $d$-dimensional point $\vec{a}$ over point $\vec{b}$.

For a set of points $Y=\{\vec{y}_0, \dots, \vec{y}_n\}$ (e.g. with $\vec{y}=\{q,t\}$), the joint ECDF becomes:
\begin{equation}\label{eq:ECDF_gen}
  \ECDF^n(\vec{y}) = \frac{1}{n} \sum_{i=1}^n \Ind{Y_i \preceq \vec{y}}
\end{equation}

\subsubsection{Of trajectories ---~EAF}

Let $Z=\{Y_0,\dots,Y_m\}$ be a set of non-dominated set of $\vec{y}$ points (e.g. $\vec{y}=\{q,t\}$).
%That is $Y_i=\{\{q_0,t_0\}, \dots \{q_r,t_r\} | q_i\leq q \land t_i\leq t \  \forall q \in Q \land \forall t \in T \}$.
Note that a set $Y\in Z$ can weakly dominate another set $\{\vec{y}\}$ of cardinality one.

The ``Empirical Attainment Function''~\cite{inp_GrunertdaFonseca2001} is defined as the ECDF of the closed set $Z$, over $\vec{y}$~\cite{a_GrunertdaFonseca2002}:
\begin{equation}\label{eq:EAF}
  \EAF^m(\vec{y}) = \frac{1}{m} \sum_{i=1}^m \Ind{Z_i \preceq \vec{y}}
\end{equation}

\subsubsection{Discretization as Histograms}

One can without loss of generality use a function $h:\reals{}^d\mapsto \reals{}^d$ which may discretize the input space,
and consider a generic set $R$ of size $r$ on which element the domination operator can be applied to compare it to any point $\vec{y}$:
\begin{equation}\label{eq:EAF_h}
  \EAF^r_h(\vec{y}) = \frac{1}{r} \sum_{i=1}^r \Ind{R_i \preceq h(\vec{y})}
\end{equation}

%We then recover the following definitions, summarized in Table~\ref{tab:glossary}:
%\begin{itemize}
%    \item The Empirical Cumulative Distribution Function (ECDF) can be defined as eq.~\ref{eq:EAF_h} with $h(\vec{y})=\vec{y}$ and $R=Y$ (e.g. a set of $\{q,t\}$ points).
%
%    \item The Empirical Attainment Function (EAF) can be defined as eq.~\ref{eq:EAF_h} with $h(\vec{y})=\vec{y}$ and $R=Z=\{Y_0,\dots,Y_m\}$, a set of ``trajectories'' (e.g. a trajectory being a set of weakly non-dominated $\{q,t\}$ points, being the best quality/time targets encountered by the optimization algorithm).
%
%    \item The Empirical Attainment Histogram (EAH) can be defined as eq.~\ref{eq:EAF_h}, using $R=Z$ and a discretization function:
%    \begin{equation}\label{eq:discretization}
%        h(\vec{y}) = \Delta^{-1} \left( \left\lfloor \Delta(\vec{y}) \right\rfloor \right)
%    \end{equation}
%
%    \item The Empirical Cumulative Distribution Histogram (ECDH) can be defined as eq.~\ref{eq:EAF_h}, using eq.~\ref{eq:discretization} and $R=Y$.
%\end{itemize}
%
%
%\begin{table}[htbp]
%\caption{Glossary.}
%\label{tab:glossary}
%\centering
%\begin{tabular}{|l|c|c|} 
%\cline{2-3}
%\multicolumn{1}{l|}{}           & Distribution (of points) & Attainment (of runs)  \\ 
%\hline
%                     Function   & ECDF         & EAF  \\ 
%\hline
%                     Histogram  & ECDH         & EAH  \\
%\hline
%\end{tabular}
%\end{table}

\subsection{2D Quality/Time Distributions}\label{sec:2DQT}

 The Quality/Time EAF can be defined as eq.~\ref{eq:EAF_h} with $h(\vec{y})=\vec{y}$ and $R=Z=\{Y_0,\dots,Y_m\}$, a set of ``attainment trajectories''.
 In our case, a trajectory is a set of weakly non-dominated $\{q,t\}$ points, being the best quality/time targets encountered by the optimization algorithm.

 The Empirical Attainment Histogram (EAH) can be defined as eq.~\ref{eq:EAF_h}, using $R=Z$ and a discretization function:
    \begin{equation}\label{eq:discretization}
        h(\vec{y}) = \Delta^{-1} \left( \left\lfloor \Delta(\vec{y}) \right\rfloor \right)
    \end{equation}

A linear discretization of the histograms can be defined by mapping on $\ints{}^+$, as an indexed space, useful for implementation:
\begin{equation}
    \Delta_\beta(\vec{y}) = \left(\vec{y}-\vec{v}\right) \frac{\beta}{l}
\end{equation}
leading to:
\begin{equation}\label{eq:disc_linear}
   h_\beta(\vec{y}) = \left\lfloor \left(\vec{y}-\vec{v}\right) \frac{\beta}{l} \right\rfloor \frac{l}{\beta} +\vec{v}
\end{equation}
where $\beta$ is the number of buckets of the histogram, $\vec{v} = \min(Y)$, the minimum corner of Y and $l=|\max(Y)-\vec{v}|$.

Similarly, a log-discretization can be defined with:
\begin{equation}
    \Delta_\beta(\vec{y}) = 
            \frac{
                \beta \cdot \log\left(
                    1 + \left(\vec{y}-\vec{v}\right)
                \right)
            }{
                \log(1+l)
            }
\end{equation}
leading to:
\begin{equation}\label{eq:disc_log}
    h_\beta(\vec{y}) =
    \exp\left(
        \left(
        \left\lfloor 
            \frac{
                \beta \cdot \log\left(
                    1 + \left(\vec{y}-\vec{v}\right)
                \right)
            }{
                \log(1+l)
            }
        \right\rfloor +1
        \right)
        \frac{\log(1+l)}{\beta}
    \right)
    -1+\vec{v}
\end{equation}

\subsection{Algorithms}

To compute the EAF, we use the algorithm of \cite{inp_Fonseca2011},
and follow the implementation provided by \cite{inc_LopezIbanez2010},
which is available in the \name{eaf}\footnote{\url{https://mlopez-ibanez.github.io/eaf/}} package fo \name{R}.
In short, the algorithm computes the level sets of attainment points $\{q,t\}$, as figured by the red lines in Figure~\ref{fig:QTEAF} (h) and (i).

%FIXME algorithm

Note that the algorithm to compute the log-EAH is adapted from \cite{inp_Knowles2005}, using the log discretization function~\ref{eq:disc_log}.

\subsection{Implementation}

The implementation takes the form of the \code{EAF} class, inheriting from the \name{Logger}
while fixing the Properties that are watched
and the Triggers to the one being necessary to the EAF computation (\code{YBest} and \code{OnImprovement}, respectively).

As we want the user to be able to compute various statistics over the EAF, we store every level set in a separated vector,
storing the metadata about the run with which they have been produced.
Each EAF is thus attached to the run, the problem instance, the dimension and the suite for which it has been produced.
This allow the user to operate any aggregation across metadata and compute any statistics.

We provide several generic statistics in the form of computation of the \code{Surface} of a given attainment level,
and the \code{Volume} of a given set of levels.
The user can use low-level classes to indicate which Nadir point they want to use, if the default worst point is not desired.

Note that the volume gives access to a first-order moment-like statistic that behave like the average,
while the surface of the median level of the EAF gives access to a robust, median-like statistics.
Second-order moment statistics may be computed, as explained in~\cite{inp_Fonseca2005}.
Additionally, it would be possible to take (or aggregate) slice(s) of the EAF
in order to recover expected runtime or expected quality distributions.

\begin{lstlisting}[language=C++,label=lst:eaf,caption={Excerpt of code used to test the behaviour of the EAF logger.}]
size_t sample_size = 10;
size_t nb_runs = 10;
suite::BBOB suite({1, 2}, {1, 2}, {10, 30});
// Instantiate the EAF logger.
logger::EAF logger;
suite.attach_logger(logger);

// Classical benchmark calls:
for(const auto& pb : suite) {
    for(size_t r = 0; r < nb_runs; ++r) {
        for(size_t s = 0; s < sample_size; ++s) {
            (*pb)(common::random::pbo::uniform(
                static_cast<size_t>(pb->meta_data().n_variables),
                static_cast<long>(s)));
        }
        pb->reset();
    }
}
// Get some EAF levels:
logger::eaf::Levels levels_at(common::OptimizationType::Minimization, {0, nb_runs/2, nb_runs-1});
auto levels = levels_at(logger);
\end{lstlisting}

The discretized EAHs (for linear scales, as defined in Section~\ref{sec:2DQT}) have been implemented in a previous work and have been ported to the new architecture.
We add support for log scales, which are usually more suited to the convex shape of the EAH.

\begin{figure} 
    \centering
    \includegraphics[width=1.0\textwidth]{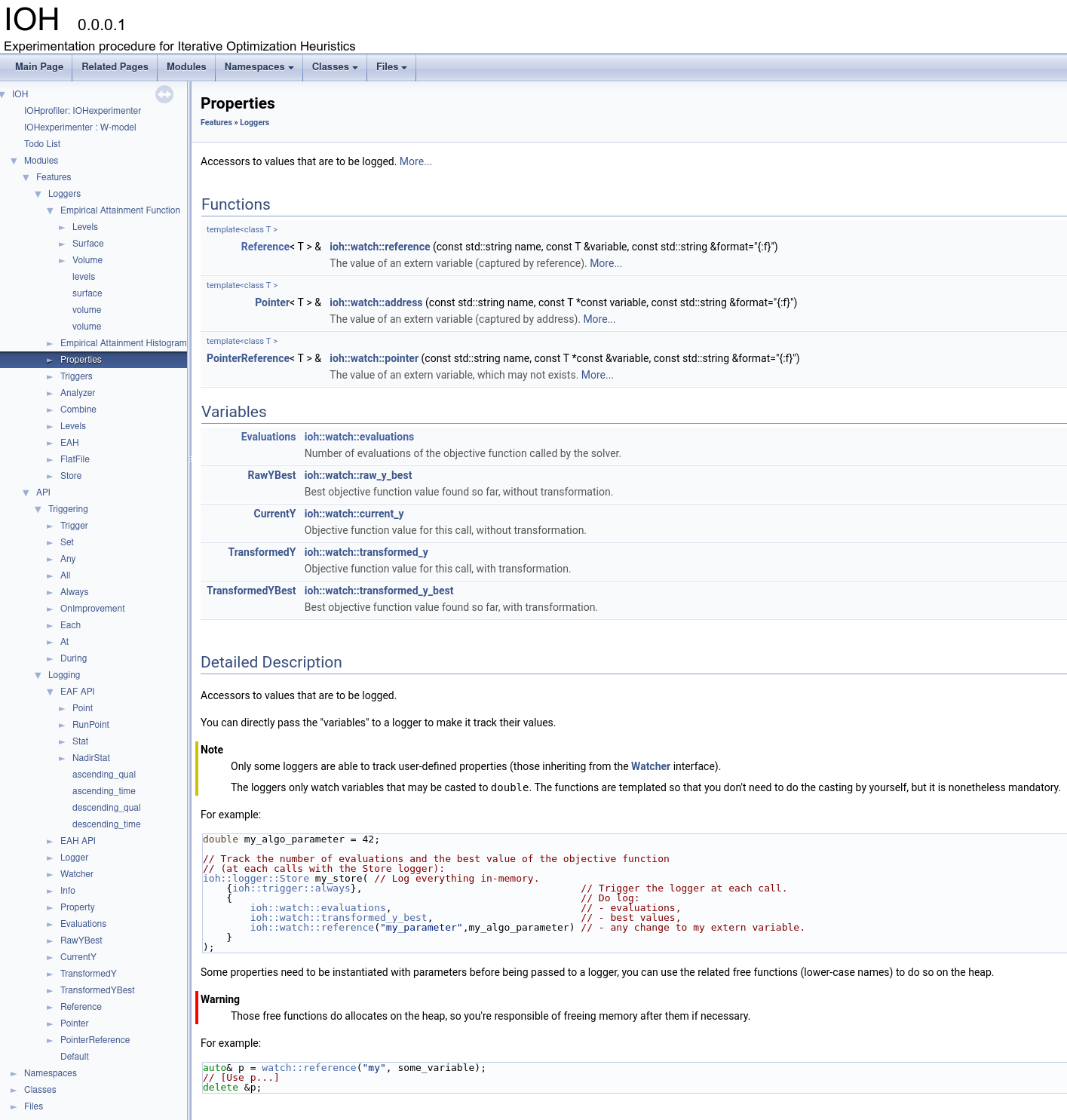}
    \caption{Screenshot of the automatically generated documentation.}
    \label{fig:doxygen} 
\end{figure}

All implemented classes are accompanied by several corresponding unit tests and an extensive documentation, as shown on Figure~\ref{fig:doxygen}.

\section{Conclusion}

In this work, we target the use of \IOH{experimenter} as a key framework for landscape-aware algorithm selection.
In order to be able to perform large-scale per-instance automated configuration,
we need to embed a powerful logger within a solver executable.
Additionally, the logger need to be able to computes various performance statistics across large benchmarks.

So as to allow for a generic performance estimation, we implement the computation of the Quality/Time Empirical Attainment Function,
a generalization of Empirical Cumulative Distribution Function for attainment trajectories of anytime stochastic algorithms.

To meet this objective, we refactor the logging system of \IOH{experimenter}, so as to obtain a highly modular and extensible architecture,
with fine-grained debugging message management.
Using this new system, we implement the EAF logger and some statistics on this distribution.
The new setup have been successfully tested on an extension of our previous work \cite{poster_coseal}.

Our implementation provides an extensive documentation and several examples in the form of unit tests.
It has eventually been merged within the \IOH{experimenter} code base\footnote{See the pull request for further details: \url{https://github.com/IOHprofiler/IOHexperimenter/pull/92}}.

%\printbibliography
\bibliographystyle{hapalike.bst}
\bibliography{QTF}

\end{document}